# AI Safeguards, Generative AI and the Pandora Box: AI Safety Measures to Protect Businesses and Personal Reputation

Prasanna Kumar
pk@businessoptima.com
Business Optima

**Abstract**

Generative AI has unleashed the power of content generation and it has also unwittingly opened the pandora box of realistic deepfake causing a number of social hazards and harm to businesses and personal reputation. The investigation & ramification of Generative AI technology across industries, the resolution & hybridization detection techniques using neural networks allows flagging of the content. Good detection techniques & flagging allow AI safety - this is the main focus of this paper. The research provides a significant method for efficiently detecting darkside problems by imposing a Temporal Consistency Learning (TCL) technique. Through pretrained Temporal Convolutional Networks (TCNs) model training and performance comparison, this paper showcases that TCN models outperforms the other approaches and achieves significant accuracy for five darkside problems. Findings highlight how important it is to take proactive measures in identification to reduce any potential risks associated with generative artificial intelligence.

**Keywords:** Generative AI, Deepfake, Temporal Consistency Learning, TCN Model, Temporal Convolution Network

## 1. Introduction

Even though generative AI has opened up a lot of creative potential, its quick development has brought about a terrifying reality, especially when it comes to deepfakes. These malicious technologies use complex algorithms to create eerily realistic audiovisual information, giving offenders the ability to make people seem as though they are talking or doing things they have never done. Dreadful outcomes, such as identity theft that destroys standing, wrongful imprisonment that wrecks lives, and even planned violence motivated by false narratives, can result from this manipulation. Such pernicious media not only destroys individual privacy but also erodes society's basic basis for truth, spreading false information and malevolent propaganda like wildfire. (Müller et al., 2024), (Mubarak et al., 2023).

Advanced deep learning techniques are used to create deepfakes, and people looking to trick and control others increasingly turn to recurrent neural networks, autoencoders, and Generative Adversarial Networks (GANs) (Gu et al., 2021). Furthermore, a number of user-friendly platforms and websites, including DeepArt, Zao, and Reface, have made it easier and easier for people to produce convincing deepfake films. These tools allow users to create hyper-realistic videos that readily deceive viewers and alter public opinion by utilizing large datasets and computer power. Because these sophisticated tools are so easily accessible, the production and distribution of deepfakes have increased dramatically, moving from being interesting side projects to serious risks in today's digital environment.

The development of generative AI has transformed media production, but when used improperly, it poses serious threats to privacy, security, and public trust. The trustworthiness of public personalities can be damaged by technologies like deepfakes, which produce convincing fake videos.. Artificial intelligence-generated material, such altered photos of celebrities, propagates false ideals and has the potential to cause defamation. Furthermore, the creation of fake news powered by AI spreads misinformation and undermines confidence in reliable news sources. Artificial intelligence-driven phishing emails that mimic well-known voices present new risks of fraud and identity theft, and bot accounts run by AI can spread false information widely and further stifle public debate. These problems underline the critical need for accurate detection tools and regulatory frameworks to prevent the adverse effects of generative AI on society.

An exploration of traditional manual detection approaches has been conducted in an effort to counter the formidable risks presented by deepfakes. To find pixel irregularities or disparities in audio-visual synchronization, these tactics include professional visual assessment and digital forensics (Rossler et al., 2019). These labor-intensive methods, however, are completely insufficient to deal with the deepfake dilemma in real-time as it is fast evolving. A purely human approach to monitoring the increasing volume of deepfake content is not only unrealistic, but also a surefire way to fail.

Here, the paper address the darkside problems epidemic with state-of-the-art machine learning methods, namely a Temporal Consistency Learning (TCL) detection strategy. The Conventional TCN Model, WavNet Model, InceptionTime Model, ConvTasNet Model, and TAGN Model are the five pre-trained Temporal Convolutional Network (TCN) models assessed. TCL highlights the dynamic differences in video sequences or the irregularities in the images that are frequently missed by traditional frame-by-frame analysis. This underscores the pressing need for sophisticated detection techniques in order to combat the threats. This study highlights how crucial it is to take preventative action in order to thwart identity theft and rebuild public confidence in the veracity of digital media.

2. Related Work

Recent years have seen a substantial amount of research in the subjects of deep fake identification and the wider ramifications of generative AI. To counter the threat of deepfakes, a variety of strategies have been investigated, from behavioral analysis to machine learning models. In a recent study, (Leporoni et al., 2024) considerably increased detection rates in big datasets by proposing a hybrid detection framework that combines temporal analysis and convolutional neural networks (CNN) to detect anomalies across video frames. Similar to this, (Kaur et al., 2024) carried out a thorough analysis of numerous deepfake detection strategies, stressing the benefits and drawbacks of current algorithms, such as frame-based and temporal-based detection techniques. They stress that further study has to concentrate on enhancing model resilience against generative models that are more and more realistic.

A novel approach for detecting video forgeries was introduced by (Harsh Vardhan et al., 2024) in April 2024. The system makes use of distinct behavioral indicators linked to human speech patterns and facial expressions. Their spatio-temporal analysis-based deepfake detection approach showed encouraging results in identifying minute changes. (Jacobsen, 2024a) investigation on the cultural and sociological ramifications of deepfake technology, notably in relation to misinformation and identity theft, was another noteworthy contribution. Their research emphasizes how important it is to have robust legal frameworks in place to prevent the abuse of generative AI.

Furthermore, for deepfake identification, (Qadir et al., 2024) investigated the merging of multimodal data, such as audio-visual cues. Their method improved the detection accuracy rate by combining static and dynamic data taken from videos. This approach has demonstrated efficacy in real-time identification cases, contributing to the expanding corpus of studies focused on enhancing detection efficiency. (Prof. Dikshendra Sarpate et al., 2024) introduced a novel approach that closely resembles the study on Temporal Consistency Learning (TCL): focusing on temporal consistency anomalies to detect deepfakes in real-time contexts.

(Jacobsen, 2024b) provided additional research on the social aspects of countering deepfakes. They looked at the psychological repercussions and societal hazards related to deepfakes, especially with regard to identity theft and reputation damage. In order to counter these expanding risks, their study emphasized the significance of combining technology solutions with public awareness. Ultimately, (A. Sathiya Priya & T. Manisha, 2024) presented a deep reinforcement learning-based method that dynamically modifies detection models in response to changing deepfake characteristics. This strategy provides a more effective and flexible way to identify deepfakes, and it works well in practical situations.

*The focus of this paper will be specifically on investigating five major issues raised by generative AI- AI-generated photoshopped images, fake news production, AI-generated phishing emails, and AI-synthesised voice fraud through functionality of Temporal Convolutional Networks (TCNs). To find out which of the five models—Conventional TCN, WavNet, InceptionTime, ConvTasNet, and Temporal Attention Graph Network (TAGN)—is compatible and efficient in countering these threats, the study will specifically assess and compare the performance of these models. The objective of this comparative analysis is to determine which of the five "dark-side" challenges has the best accuracy. This will help to highlight the advantages and disadvantages of each model in the fight against the improper use of generative AI.*

## 3. Temporal Convolutional Networks

A specific deep learning architecture called a Temporal Convolutional Network (TCN) is made to efficiently handle sequential data. TCNs use convolutional layers to more effectively capture temporal dependencies in data, in contrast to conventional recurrent neural networks (RNNs), which process inputs sequentially. Convolutional layers, as opposed to recurrent layers, are used in models such as RNNs and Long Short-Term Memory (LSTM) networks to enable TCNs to assess data over a wider range of temporal contexts without the computational load that comes with sequential processing. This is made possible by the use of dilated convolutions, which give TCNs a distinct advantage for tasks like time-series analysis, audio processing, and video classification, including deepfake detection. TCNs can expand their receptive field and analyze data from multiple time steps at once (Bai et al., 2018a).

### 3.1 Architecture of TCNs

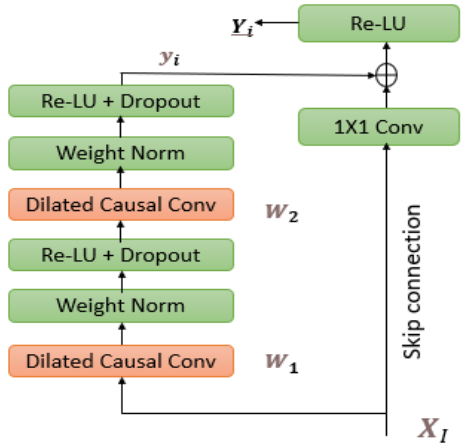

Figure 1: TCN residual block. An 1x1 convolution is added when residual input and output have different dimensions.

As seen in Figure 1, a TCN's architecture is distinguished by the usage of residual blocks. Dilated causal convolutions are incorporated into every residual block to aid in the learning of complex temporal patterns across frames. To add non-linearity, the rectified linear unit (ReLU) activation function is utilized, and weight normalization is used to stabilize training and enhance model performance. Additionally, each dilated convolution layer is merged with spatial dropout, which periodically omits entire channels during training to reduce overfitting. (Bai et al., 2018a). The equation for a dilated convolution can be expressed as follows:

$$y[t] = \sum_{k=a}^{k-1} x[t - d.k] . w[k] \quad [1]$$

where the output at time t is represented by $y[t]$, the input signal by $x[t]$, the kernel at location k is indicated by $w[k]$, the dilation factor by d and the kernel size by **q** K. This formula demonstrates how TCNs evaluate inputs at different temporal scales, improving their capacity to identify minute temporal anomalies that are frequently found in altered video content.

In the above equation, the dilation factor $d$ determines how many times the network is able to analyze temporal patterns. For instance, if the dilation factor $d$ is larger, the network can cover a much longer time window in fewer layers and thereby significantly increasing the receptive field. Generally, the overall receptive field R of the network can be written as:

$$R = 1 + (K - 1) . \sum_{l=0}^{L-1} d_l \quad [2]$$

Where $L$ is the total number of layers in the TCN, and $d_l$ is the dilation factor for layer $l$

To introduce non-linearity, each residual block applies Rectified Linear Unit (ReLU) activation function after the convolution operation.

$$y`(t) = max(0, y(t)) \quad [3]$$

Here $y`(t)$ is the output after applying ReLu and $y(t)$ is the output of the convolution operation. A spatial dropout is introduced with each layer to reduce overfitting. Weight Normalization applied to convolutional filters to enhance model performance and make model stable. L2 norm technique is used here.

$$\widehat{w} = \frac{w}{|w|} \quad [4]$$

Where $|w|$ represents the L2 norm of weight vector. TCNs' parallel processing powers greatly accelerate training and inference as compared to RNNs, whose sequential design results in slower performance. Additionally, residual connections increase gradient flow through the network by resolving problems with disappearing gradients (Sheng et al., 2024). With this approach, TCNs can accurately discern between real and deepfake videos by continuously analyzing video frames for temporal irregularities (Pelletier et al., 2019). TCNs, due to their convolutional nature can process all time steps simultaneously on the network, this allows for much faster training and inference.

### 3.2 Utility of Pretrained Models

Identification techniques can be made much more accurate and resilient by using pre-trained TCN models. Large, diverse datasets are commonly used to train these models, allowing them to acquire generalized features that may be applied to a range of applications. We can take advantage of these pretrained models' prior knowledge to enhance their ability to recognize faked videos by honing them on certain deepfake datasets. While some pretrained models may be better at seeing specific artifacts created during the deepfake production process, others might be better at spotting temporal differences (Kaur et al., 2024). One can compute the final forecast as follows:

$$\widehat{y_{final}} = argmax\left(\frac{1}{M}\sum_{m=1}^{m}\widehat{y_m}\right) \quad [5]$$

Where $\widehat{y_{final}}$ is the final predicted output and $\widehat{y_m}$ is the predicted output from model $m$.

Furthermore, using several pretrained models enables cross-validation of the findings, improving the detection system's dependability. These models' combined insights can offer a more thorough comprehension of the subtleties related to deepfakes, which could ultimately result in the creation of improved and more effective detection methods (Sultan & Ibrahim, n.d.). This strategy is in line with deep learning best practices, which show that using several models frequently results in better performance than single-model architectures.

### 4. Methodology

A methodical procedure for training and validating algorithms in a machine learning setting is described in figure 2.

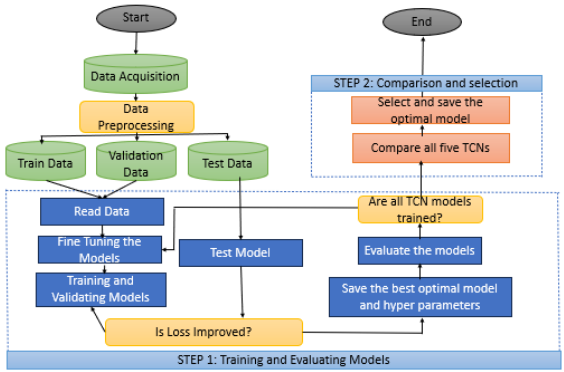

Figure 2: Flowchart of the proposed solution

Data collection is the first step, and then preprocessing is done to get the data ready for analysis. Subsets of the dataset are designated as train, test and validation data. The initial phase involves reading the data, validating the models and fine-tuning different models to ensure exceptional results. The models undergo evaluation after training to see if the loss is decreased. Comparing each trained model to determine which is best based on performance indicators for the compatible darkside problem is the second stage. This thorough procedure guarantees that the top performing model and its hyperparameters remain unchanged, allowing for efficient deployment in real-world applications. The solution for the proposed system can be understood through the below algorithm.

| Algorithm: Darkside Solutions |
|---|
| **Require:** Dataset D<br>**Output:** Trained model for classifying instances d∈D |
| 1. Data Collection<br>Gather instances for each dark-side problem:<br>- **Deepfakes:** Collect videos and their corresponding frames.<br>- **Photoshopped Images:** Collect authentic and manipulated images.<br>- **Fake News:** Gather real news articles and fake news articles.<br>- **Phishing Emails:** Collect legitimate and phishing email samples.<br>- **Synthesized Voice:** Record real voices and AI-generated synthetic voices.<br><br>2. Data Preprocessing<br>Image Resizing, Text Processing-Tokenization, Data Augmentation, Normalization, Text Vectorization, etc.<br><br>3. Data Splitting<br>Training Data: 70%<br>Validation Data: 30%<br><br>4. Model Training<br>Import a set of models M = {TCN, WaveNet, InceptionTime, ConvTasNet, TAGN}.<br>For each model m ∈ M do<br>    Initialize learning rate α = 0.001<br>    for epochs = 1 to 30 do<br>        foreach mini batch (X_train, Y_train) do<br>          Update parameters of the model<br>          if validation error does not improve for 5 consecutive epochs then<br>            α = α × 0.1   // Reduce learning rate |

> 5. Model Evaluation
> Calculate evaluation metrics in a set of metrics T: Accuracy and Loss, Precision, Recall , F1 Score, Confusion Matrix
>
> 6. Match & Summarize
> Compare performance metrics across all models and select the best-performing model for each dark-side problem.
> For problem in deepfake problems:
>     If T(Model[0]) is max:
>         CTCN is Solution
>     Else if T(Model[1]) is max:
>         Wavenet is Solution
>     Else if T(Model[2]) is max:
>         Solutionis Solution
>     Else if T(Model[3]) is max:
>         InceptionTime is Solution
>     Else T(Model[4]) is max:
>         TAGN is Solution
> Print: Darkside Problem, Overlay Warning Messages on the Fake Content, and relevant Media.
>
> 7. Register the Model
> Register the top model by using tensorflow SavedModel
> Store model metadata such as performance metrics (precision score, f1 score, etc), architecture name (CTCN, WaveNet, etc)

### 4.1 Dataset

To address a variety of generative AI-related dark-side issues, several datasets were used in this study. Over 100,000 annotated movies, including both authentic and deepfake content, are available in the Deepfake Detection Challenge (DFDC) dataset from Kaggle. In order to ensure robust model training, Facebook worked with AI researchers to create this dataset, which is a benchmark for improving deepfake detection algorithms. It includes a variety of race, gender, lighting, and environmental characteristics. Furthermore, millions of brief statements from several sources are sorted by veracity in the LIAR dataset, which is used to identify false news creation using AI. With a large collection of authentic email exchanges, the Enron Email Dataset facilitates the analysis of artificial intelligence (AI)-generated phishing emails. The VCTK Corpus provides a wide variety of voice recordings and is used to look into AI-generated voices in fraudulent operations. Lastly, data from several accounts recognized for bot-like activity is included in the Twitter Bot Dataset, which is used to research AI-controlled bot accounts that disseminate false information. When combined, these datasets offer a strong foundation for training models that can successfully address the difficulties presented by generative AI. The overview of datasets is explained in table 1.

| Datasets | Count | Description |
| --- | --- | --- |
| Deepfake Detection Challenge (DFDC) | 124,645 | Original unaltered videos (97,919) and manipulated videos (26,728) using deepfake technology. |
| LIAR | 12,836 | Collection of 12,836 short statements from various sources categorized by truthfulness (true, mostly true, half true, mostly false, false). |
| Enron Email | 500,000 | Compilation of real emails from the Enron Corporation, including a wide range of communications from employees. |
| VCTK Corpus | 44,000 | Contains 44,000 recordings of English speech from 109 speakers with various accents, used for synthesizing voices. |
| Twitter Bot | 1,000,000+ | Dataset of tweets from known bot accounts, comprising over 1 million tweets to study misinformation spread. |

Table 1: Overview of Datasets

### 4.2 Data Preprocessing

I used crucial preprocessing techniques to improve the quality of the input data and maximize model performance across a range of dark-side challenges. In order to ensure a range of [0, 1] for effective model convergence, I divided the pixel values for image datasets by 255.0. Among the methods used to replicate realistic fluctuations in the data were random rotation, shearing transformations, zoom adjustments, horizontal flipping, brightness modifications, and channel shifting. I used vectorization using TF-IDF or word embeddings, stop word removal, stemming/lemmatization, and tokenization for textual datasets.

I used upsampling to precisely solve the problem of data imbalance in the Deepfake Detection Challenge (DFDC) dataset. The dataset was initially composed of 73,254 real photographs and 20,699 false images, which could have caused bias in the training of the model. I equalized the sample counts to 73,154 for both classes by upsampling the minority class, which enhanced the model's capacity to learn from both real and fake information. Additionally, I reduced the batch size to 16 and resized images to 64x64 pixels for efficient training while managing memory constraints. These preprocessing techniques greatly increased the quality of the datasets and boosted the models' capacity to identify different problems connected to generative AI.

|  | Before Upsampling | | After Upsampling | |
|---|---|---|---|---|
|  | Real | Fake | Real | Fake |
| Train | 73154 | 20699 | 73154 | 73154 |
| Validation | 24765 | 6029 | 24765 | 24765 |

Table 2: Comparison of real data with upsampled data for DFDC dataset

## 4.3 TCNs Pretrained model training and Comparison

For this research, I opted for Conventional Temporal Convolutional Network (CTCN), WaveNet, InceptionTime, ConvTasNet and Topology Adaptive Graph Networks (TAGN) as the five advanced models for detection of ramification. Each one of these models has a distinct function in identifying structure irregularities, which are essential for detection. A thorough description of each model, including its features, advantages, and application in the script, is provided below. It also includes any pertinent formulas for identifying deepfakes.

### 4.3.1 Conventional Temporal Convolutional Network (CTCN)

CTCN are the extension of TCN specifically designed to handle sequential data and capture temporal dependencies effectively. They are used in time series and video frame analysis tasks, as CTCN can analyze sequential data efficiently because of its ability to detect long-range temporal dependencies via dilated convolutions. TCN detects artifacts and glitches suggestive of manipulations that human observers can miss in the backdrop of deepfake detection by analyzing temporal anomalies between video frames. In this approach, dilated convolution layers are used by TCN to evaluate video frame sequences, enabling the model to capture associations between frames at various time intervals. To assess video frame sequences in this method, I used dilated convolution layers, which allowed the model to identify correlations between frames at different times. Three convolutional layers make up the model. The equation can be used to illustrate TCN's primary function.:

$$y(t) = \sum_{k=0}^{k-1} x(t - k.d) . \omega(k) \quad [6]$$

Here, the model scans back at $k$ prior frames for each time t, where $k$ varies from 0 to $K - 1$. The model's appearance is determined by multiplying each frame by a corresponding weight, $w(k)$, and the dilation factor, d (Bai et al., 2018b). The output $y(t)$ is the current point in time t is produced by adding up the values, which helps to find patterns over time and spot discrepancies that could be signs of deepfakes.

### 4.3.2 Wavenet

High-fidelity audio production is a well-known benefit of the dilated convolutions-based WaveNet architecture, which has also been effectively used to a number of sequential data analysis tasks. WaveNet's more flexible and hierarchical structure allows for numerous layers of dilated convolutions with various dilation rates, in contrast to Temporal Convolutional Networks (TCN), which typically concentrate on capturing temporal correlations through set dilation factors. Because of this architecture, the network can better predict complicated interactions over a range of time periods and discover complex patterns in the data. Using WaveNet's flexible architecture, I improved the receptive field in my implementation for AI-generated speech recognition by adding a dilation rate of two, which helped the model recognize complex patterns. WaveNet efficiently interprets data for fraudulent activity detection by looking at pixel-level correlations between audio waveforms, which enables it to capture both short- and long-range interactions.

$$y(t) = \sum_{k=0}^{K-1} w_k . x(t - d.k) \quad [7]$$

where $y(t)$ represents the output at time $t$, $x$ denotes the input sequence, and $w_k$ are the learned weights. Through this iterative process, WaveNet effectively identifies abnormalities in the transitions. WaveNet's fundamental function can be expressed similarly as in the TCN, where the dilation factor $d$ allows for adjustable backtracking. WaveNet learns to identify abnormalities in transitions through this iterative process, which is essential for identification (Oord et al., 2016).

### 4.3.3 InceptionTime

InceptionTime successfully extracts features from sequential data using a concurrent convolution method with different kernel sizes. Using parallel convolutions, it aims to capture both minute details and more larger patterns. InceptionTime increases detection accuracy by effectively identifying both minor and big anomalies in deepfake detection by analyzing video frames at different sizes. We propose an approach that enables the simultaneous extraction of features at several resolutions. Using kernel sizes of (1 x 1) and (3 x 3), I created InceptionTime to process video frames over multiple convolution routes. This architecture has two branches that each have these kernel sizes, which enhance the model's capacity to recognize and learn from a wide range of features. The process can be stated as:

$$y = f_1(X) + f_3(X) + f_5(X) \quad [8]$$

When the input video frames, X, are processed using the following convolution functions: $f_1$, $f_3$, and $f_5$. The outputs of each convolution, which capture different feature scales, are added together to get the final result ($y$) (Fawaz et al., 2020).

### 4.3.4 ConvTasNet

ConvTasNet is most known for its remarkable capacity to distinguish between audio signals, but it can also be modified to recognize important components in image data. In my solution, I kept a TCN-like architecture that was optimized for photos while modifying ConvTasNet to concentrate on pertinent pixel patterns for identifying changed regions in video data. This model excels at filtering out noise while preserving relevant information, allowing it to successfully isolate crucial patterns necessary for deepfake identification. By processing sequences of video frames, ConvTasNet boosts the model's potential to spot inconsistencies and anomalies in visual content, helping significantly to accurate deepfake recognition.

$$Y = XW + N \quad [9]$$

Where $Y$ represents the output, $X$ is the input feature matrix, $W$ is the weight matrix, and $N$ denotes noise (Luo & Mesgarani, 2019).

### 4.3.5 Topology Adaptive Graph Networks (TAGN)

In order to comprehend spatial relationships between data points, it employs a graph-based technique and examines pixels as nodes in a graph. In this system, we treated video frames as graphs, where edges stand in for spatial links and each pixel is a node. This capacity to record intricate geometric relationships is also helpful for deepfake detection, as TAGN looks for structural irregularities in facial features by examining the spatial correlations between important visual characteristics in video frames. The model incorporates standard Conv2D layers, which are developed from traditional graph networks. The graph convolution formula describes the action of TAGN as follows:

$$Y = \sigma(D^{-1/2} A D^{-1/2} XW) \quad [10]$$

where D is a measure of node connection and A is the adjacency matrix that shows relationships between pixels. The normalized adjacency matrix $D^{-1/2} A D^{-1/2}$ is multiplied by the input feature matrix $X$ and weight matrix $W$ to perform the convolution. An activation function $\sigma$ is then applied to yield the output $Y$ (Monti et al., 2016).

### 4.4 Fine-Tuning Hyperparameters

To improve performance, I used a methodical strategy to deploy multiple models in this research.

| Model | Layers | Batch Size | Epochs | LR | Optimizer |
|---|---|---|---|---|---|
| TCN Model | 28 | 8, 16, 32 | 50 | 0.0001, 0.0005, 0.00001 | Adam, SGD |
| WavNet Model | 58 | | | | |
| InceptionTime Model | 58 | | | | |
| ConvTasNet Model | 38 | | | | |
| TAGN Model | 38 | | | | |

Table 3: Hyperparameter Tuning

The Temporal Convolutional Network (TCN) model was first set up with 28 layers and trained the model over 50 epochs, experimenting with batch sizes of 8, 16, and 32. I used optimizers like Adam and SGD to test learning rate values of 0.0001, 0.0005, and 0.00001. Subsequently, the WavNet model, which has 58 layers in its structure, was the next thing I addressed. For consistency in comparison, the InceptionTime model likewise had 58 layers. Next, I built the ConvTasNet model, which had 38 layers, and lastly, the TAGN model, which also had 38 layers. This organized arrangement among several models was essential to assessing their efficacy and accomplishing the research goals.

## 5. Results

This section includes a graphical depiction of the performance metrics and metrics of the five models that the algorithm assessed.

| Model | Precision | Recall | F1 Score |
|---|---|---|---|
| CTCN Model | 0.9514 | 0.9712 | 0.9613 |
| WavNet Model | 0.9601 | 0.9721 | 0.957 |
| InceptionTime Model | 0.9634 | 0.9539 | 0.9586 |
| ConvTasNet Model | 0.9471 | 0.9345 | 0.9408 |
| TAGN Model | 0.9251 | 0.9441 | 0.9346 |

Table 4: Performance Metrics for All Model

The precision, recall, and F1 score for all of the five models we employed in these studies are compiled in Table 4. These metrics are essential for assessing how well the models detect irregularities with the least amount of false positives and false negatives.

| Model | Train Acc | Val Acc | Test Acc |
|---|---|---|---|
| CTCN Model | 0.9918 | 0.9812 | 0.9865 |
| WavNet Model | 0.9902 | 0.9792 | 0.9847 |
| InceptionTime Model | 0.9634 | 0.9639 | 0.9637 |
| ConvTasNet Model | 0.9787 | 0.9731 | 0.9759 |
| TAGN Model | 0.9693 | 0.9649 | 0.9671 |

Table 5(a): Accuracy of All Models

| Model | Train Loss | Val Loss | Test Loss |
|---|---|---|---|
| CTCN Model | 0.0767 | 0.1115 | 0.0941 |
| WavNet Model | 0.0802 | 0.0997 | 0.0889 |
| InceptionTime Model | 0.1646 | 0.1521 | 0.0155 |
| ConvTasNet Model | 0.1183 | 0.1322 | 0.1212 |
| TAGN Model | 0.1452 | 0.1403 | 0.1232 |

Table 5(b): Loss of All Models

The accuracy and loss metrics for testing, validation, and training for each of the five models are shown in Table 5(a) and Table 5(b), respectively. These measures aid in evaluating the models' performance at each stage of training as well as their capacity for generalization on untested data. The Conventional TCN Model has proven to perform the best out of the five models for deepfake detection that were examined by scoring accuracy of 0.9918 and 0.9812 and 0.0767 and 0.1115 of train and validation shown in figure 3(a) and 3(b), respectively. In terms of model applicability, WaveNet is excellent for audio tasks, particularly for recognizing AI-generated voices in fraudulent operations, whereas InceptionTime excels in processing phishing emails, effectively catching intricate patterns in sequential data. ConvTasNet is a powerful tool for visual data separation since it can recognize altered regions in video footage, while TAGN is specifically made for fake news detection.

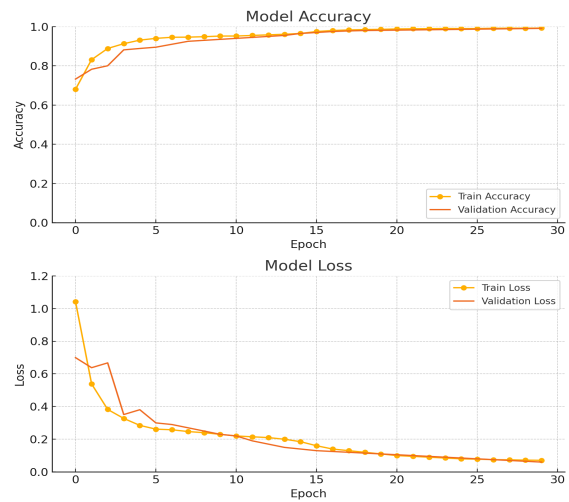

Figure 3(a): Accuracy graph for CTCN, Figure 3(b): Loss graph for CTCN

## 5.1 Summary of Findings

| No | DSP | Model | Score | Dataset |
|---|---|---|---|---|
| 1 | AI-Generated Photoshopped Celebrities or DeepFake | CTCN | 0.9918 | DFDC |
| 2 | Fake News Generation via AI | TAGN | 0.9700 | LIAR |
| 3 | AI-Generated Phishing Emails | IceptionTime | 0.9634 | Enron Email |
| 4 | AI-Synthesized Voice for Fraudulent Activities | Wavenet | 0.9902 | VCTK Corpus |
| 5 | AI-Controlled Bot Accounts for Misinformation Spread | ConTasnet | 0.9721 | Twitter Bot |

Table 6: Selection and compatibility of models

## 6. Conclusion

Generative AI can lead to misinformation, invasions of privacy, and manipulation of public opinion, all of which can have negative effects in the social, political, and economic realms. We provide a unique solution by hybridization approaches & utilizing the TCLs that use Temporal Convolutional Networks, especially the Conventional TCN model to precisely evaluate and classify video information. We offer efficient detection and mitigation of risks applicable to a number of sectors, including the financial domain & security vertical, where the veracity of visual content is crucial, making this research approach indispensable. Furthermore, the score-led if-else approach offers rule based personalization and safeguard against the unlawful use, which is a growing threat to personal privacy in the digital age. Improving detection algorithms, strengthening defenses against adversarial attacks, and developing AI governance frameworks to guarantee the moral application of these technologies ought to be the main goals of future research. Putting detection of GenAI content and flagging contents for Safe AI safety are key best practices in creating accountability and trust in AI systems. This will ensure businesses and individual rights are protected while leveraging the power of generative AI technology for content creation and consumption.

## 5. References

A. Sathiya Priya & T. Manisha. (2024). CNN and RNN using Deepfake detection. *International Journal of Science and Research Archive*, *11*(2), 613–618. https://doi.org/10.30574/ijsra.2024.11.2.0460

Bai, S., Kolter, J. Z., & Koltun, V. (2018a). *An Empirical Evaluation of Generic Convolutional and Recurrent Networks for Sequence Modeling* (Version 2). arXiv. https://doi.org/10.48550/ARXIV.1803.01271

Bai, S., Kolter, J. Z., & Koltun, V. (2018b). *An Empirical Evaluation of Generic Convolutional and Recurrent Networks for Sequence Modeling* (No. arXiv:1803.01271). arXiv. http://arxiv.org/abs/1803.01271

Campbell, C., Plangger, K., Sands, S., Kietzmann, J., & Bates, K. (2022). How Deepfakes and Artificial Intelligence Could Reshape the Advertising Industry: The Coming Reality of AI Fakes and Their Potential Impact on Consumer Behavior. *Journal of Advertising Research*, *62*(3), 241–251. https://doi.org/10.2501/JAR-2022-017

Fawaz, H. I., Lucas, B., Forestier, G., Pelletier, C., Schmidt, D. F., Weber, J., Webb, G. I., Idoumghar, L., Muller, P.-A., & Petitjean, F. (2020). InceptionTime: Finding AlexNet for Time Series Classification. *Data Mining and Knowledge Discovery*, *34*(6), 1936–1962. https://doi.org/10.1007/s10618-020-00710-y

Gu, Z., Chen, Y., Yao, T., Ding, S., Li, J., Huang, F., & Ma, L. (2021). *Spatiotemporal Inconsistency Learning for DeepFake Video Detection* (No. arXiv:2109.01860). arXiv. http://arxiv.org/abs/2109.01860

Harsh Vardhan, Naman Varshney, Manoj Kiran R, Pradeep R, & Dr. Latha N.R. (2024). Deep Fake Video Detection. *International Research Journal on Advanced Engineering Hub (IRJAEH)*, *2*(04), 830–835. https://doi.org/10.47392/IRJAEH.2024.0117

Jacobsen, B. N. (2024a). Deepfakes and the promise of algorithmic detectability. *European Journal of Cultural Studies*, 13675494241240028. https://doi.org/10.1177/13675494241240028

Jacobsen, B. N. (2024b). Deepfakes and the promise of algorithmic detectability. *European

Kaur, A., Noori Hoshyar, A., Saikrishna, V., Firmin, S., & Xia, F. (2024). Deepfake video detection: Challenges and opportunities. *Artificial Intelligence Review*, *57*(6), 159. https://doi.org/10.1007/s10462-024-10810-6

Leporoni, G., Maiano, L., Papa, L., & Amerini, I. (2024). A guided-based approach for deepfake detection: RGB-depth integration via features fusion. *Pattern Recognition Letters*, *181*, 99–105. https://doi.org/10.1016/j.patrec.2024.03.025

Luo, Y., & Mesgarani, N. (2019). Conv-TasNet: Surpassing Ideal Time-Frequency Magnitude Masking for Speech Separation. *IEEE/ACM Transactions on Audio, Speech, and Language Processing*, *27*(8), 1256–1266. https://doi.org/10.1109/TASLP.2019.2915167

Monti, F., Boscaini, D., Masci, J., Rodolà, E., Svoboda, J., & Bronstein, M. M. (2016). *Geometric deep learning on graphs and manifolds using mixture model CNNs* (No. arXiv:1611.08402). arXiv. http://arxiv.org/abs/1611.08402

Mubarak, R., Alsboui, T., Alshaikh, O., Inuwa-Dutse, I., Khan, S., & Parkinson, S. (2023). A Survey on the Detection and Impacts of Deepfakes in Visual, Audio, and Textual Formats. *IEEE Access*, *11*, 144497–144529. https://doi.org/10.1109/ACCESS.2023.3344653

Müller, N. M., Czempin, P., Dieckmann, F., Froghyar, A., & Böttinger, K. (2024). *Does Audio Deepfake Detection Generalize?* (No. arXiv:2203.16263). arXiv. http://arxiv.org/abs/2203.16263

Oord, A. van den, Dieleman, S., Zen, H., Simonyan, K., Vinyals, O., Graves, A., Kalchbrenner, N., Senior, A., & Kavukcuoglu, K. (2016). *WaveNet: A Generative Model for Raw Audio* (No. arXiv:1609.03499). arXiv. http://arxiv.org/abs/1609.03499

Pelletier, C., Webb, G., & Petitjean, F. (2019). Temporal Convolutional Neural Network for the Classification of Satellite Image Time Series. *Remote Sensing*, *11*(5), 523. https://doi.org/10.3390/rs11050523

Prof. Dikshendra Sarpate, Abrar Mungi, Shreyash Borkar, Shravani Mane, & Kawnain Shaikh. (2024). A Deep Approach to Deep Fake Detection. *International Journal of Scientific Research in Science, Engineering and Technology*, *11*(2), 530–534. https://doi.org/10.32628/IJSRSET2411274

Qadir, A., Mahum, R., El-Meligy, M. A., Ragab, A. E., AlSalman, A., & Awais, M. (2024). An efficient deepfake video detection using robust deep learning. *Heliyon*, *10*(5), e25757. https://doi.org/10.1016/j.heliyon.2024.e25757

Rossler, A., Cozzolino, D., Verdoliva, L., Riess, C., Thies, J., & Niessner, M. (2019). FaceForensics++: Learning to Detect Manipulated Facial Images. *2019 IEEE/CVF International Conference on Computer Vision (ICCV)*, 1–11. https://doi.org/10.1109/ICCV.2019.00009

Sheng, Z., Cao, Y., Yang, Y., Feng, Z.-K., Shi, K., Huang, T., & Wen, S. (2024). Residual Temporal Convolutional Network With Dual Attention Mechanism for Multilead-Time Interpretable Runoff Forecasting. *IEEE Transactions on Neural Networks and Learning Systems*, 1–15. https://doi.org/10.1109/TNNLS.2024.3411166

Sultan, D. A., & Ibrahim, D. L. M. (n.d.). *A Comprehensive Survey on Deepfake Detection Techniques*.

*Journal of Cultural Studies*, 13675494241240028. https://doi.org/10.1177/13675494241240028